\documentclass[10pt,twocolumn,letterpaper]{article}

\usepackage{cvpr}
\usepackage{times}
\usepackage{epsfig}
\usepackage{graphicx}
\usepackage{amsmath}
\usepackage{amssymb}
\usepackage{array}

\usepackage{multirow}
\usepackage{fancyhdr}
\usepackage{hyperref}
\usepackage{booktabs}
\usepackage{caption}

\usepackage{float}

\newcommand{\PreserveBackslash}[1]{\let\temp=\\#1\let\\=\temp}
\newcolumntype{C}[1]{>{\PreserveBackslash\centering}p{#1}}
\newcolumntype{R}[1]{>{\PreserveBackslash\raggedleft}p{#1}}
\newcolumntype{L}[1]{>{\PreserveBackslash\raggedright}p{#1}}

\cvprfinalcopy 

\def\cvprPaperID{****} 

\begin{document}

\title{Technical Report for Soccernet 2023 - Dense Video Captioning}  
\author{
Zheng Ruan\textsuperscript{1,2}
\quad
Ruixuan Liu\textsuperscript{1}
\quad
Shimin Chen\textsuperscript{1}
\quad
Mengying Zhou\textsuperscript{1,2}
\\
Xinquan Yang\textsuperscript{1}
\quad
Wei Li\textsuperscript{1}
\quad 
Chen Chen\textsuperscript{1}
\quad
Wei Shen\textsuperscript{1}
\\
\textsuperscript{1}OPPO Research Institute
\quad
\textsuperscript{2}Tongji University
\\
{\tt\small \{liuruixuan, chenshimin1, yangxinquan, liwei19, chenchen, shenwei12\}@oppo.com}
\\
{\tt\small \{2130730, 2130904\}@tongji.edu.cn}
}

 \twocolumn[{%
 \renewcommand\twocolumn[1][]{#1}%
 \maketitle
 \begin{center}
     \centering
     \includegraphics[height=6cm,width=13cm]{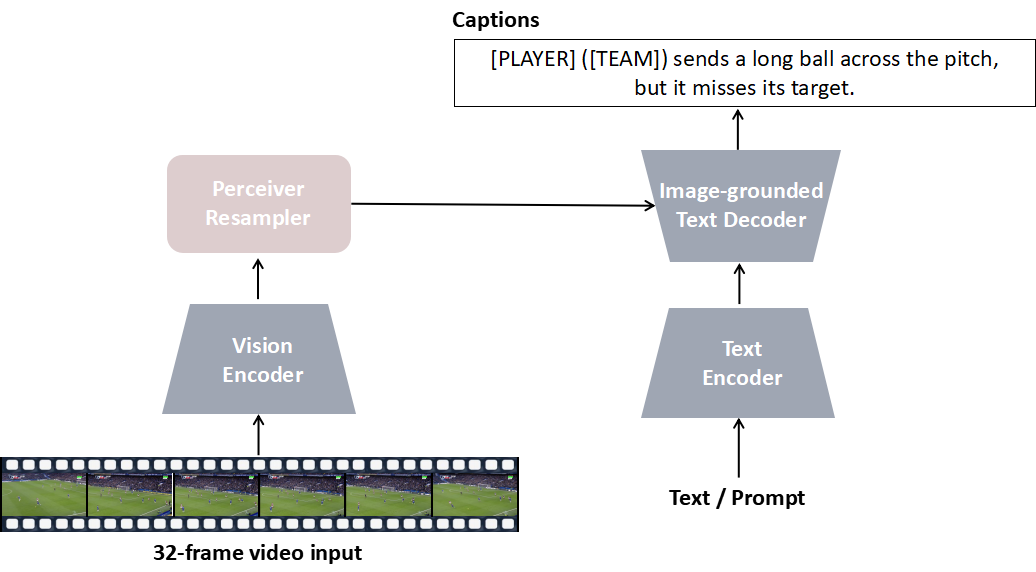}
     \captionof{figure}{Overview of our framework. Given a 32-frame video input and a prompt, our framework can generate a caption that describes the soccer actions.}
     \label{figure: Overview of our framework}
\end{center}%
 }]

\maketitle
\thispagestyle{empty}

\setlength{\parindent}{0pt}
\section{Method}
In the task of dense video captioning of Soccernet dataset, we propose to generate a video caption of each soccer action and locate the timestamp of the caption. Firstly, we apply Blip~\cite{li2022blip} as our video caption framework to generate video captions. Then we locate the timestamp by using (1) multi-size sliding windows (2) temporal proposal generation and (3) proposal classification.

\subsection{Video Caption} \label{sec:fea}
For the video caption framework, we modify Blip to fit video input and add a perceiver resampler~\cite{alayrac2022flamingo} to improve the performance. The overview of our framework is shown in Figure \ref{figure: Overview of our framework}.
For the vision encoder, we use a visual transformer~\cite{dosovitskiy2020image} to extract visual features. To adapt the video input, we add a temporal embedding layer to the vision encoder to learn the temporal information. Besides, we add a perceiver resampler block after the encoder. It set a fixed-length learned latent query to extract unified visual representations. We employ BERT~\cite{devlin2018bert} as the text encoder to extract language features. For the image-grounded text decoder, it combines the features from perceiver resampler and text encoder to generate captions.

We also train other existing methods, such as Flamingo~\cite{alayrac2022flamingo}, Blip-2~\cite{li2023blip}, etc. The experiment results are shown in Section 2.

\subsection{Localization}
\subsubsection{Central De-duplication and Threshold Filtering}
The generation of captions in the neighboring timestamps may be similar, it is necessary to filter out the duplicate sentences. Therefore, we propose a central de-duplication method. Specifically, we first calculate the number of duplicate sentences and then select the middle sentence as the output of this timestamp. In addition, we use a threshold to filter the low-quality caption. When the confidence of the output caption is lower than the set threshold, the caption will be removed.

\begin{table*}[h]
    \centering
    \caption{Comparison with different ensemble methods.}
    \label{tab:Ensemble}
    \begin{tabular}{@{}c|c|c@{}}
    \toprule
    Method                     & CIDEr           & Meteor          \\ \midrule
    Flamingo                       & 23.028          & 20.908          \\
    Blip(ViT-Base)                           & 19.957          & 21.979          \\
    Blip(ViT-Large)                           & \textbf{23.107}          & 21.131          \\
    Blip-2(ViT-Large)                          & 22.427          & 21.558          \\
    Blip(ViT-Base) + Flamingo*0.7      & 19.972 & 21.979 
    \\ 
    Blip(ViT-Base) + Blip(ViT-Large)*0.82   & 20.927 & 21.993
    \\ 
    Blip(ViT-Base) + Blip-2*0.95(ViT-Large)       & 20.693 & 21.982 
    \\
    Blip(ViT-Base) + Blip(ViT-Large)*0.85 + Blip-2(ViT-Large)*0.95      & 21.630 & \textbf{21.995} 
    \\
    \bottomrule
    \end{tabular}
\end{table*}

\subsubsection{Background Information Filtering}
To better distinguish foreground and background information, we train a VideoCLIP~\cite{xu2021videoclip} to perform information filtering. Specifically, we train three different VideoCLIP, i.e., only applying patch embedding or position embedding, and both patch and position embedding. After obtaining the caption from the VideoCLIP, we perform a simple weighted mean using the score of the caption and filter the result by a threshold. By this means, the background information can be greatly filtered.

\subsection{Ensemble}
\subsubsection{Caption Confidence}
We generate the captions of each timestamp with the confidence by three prediction models, i.e., BLIP\cite{li2022blip}, BLIP-2\cite{li2023blip} and Flamingo~\cite{alayrac2022flamingo}. Considering that the confidence of each caption generated by different models has a gap, we assign different weights for the models to calculate a new score before the ensemble. For each timestamp of video in these three models, we select the caption with the top-1 score as the final result.

\subsubsection{Vicuna-7b}
Using the confidence of each caption for ensemble can not capture the relationship between different captions. Take the following two sentences as examples, "[COACH] has decided to introduce fresh legs, with [PLAYER] ([TEAM]) replacing [PLAYER]" and "[TEAM] will have a chance to score from a corner kick. a substitution has been made. [PLAYER] is replaced by [PLAYER] ([TEAM])" have similar contexts, but there are differences in some descriptions (corner kick). The ensemble method of selecting the highest confidence sentence will discard these different descriptions. An optimal method is to combine these captions and refine the sentence. Inspired by the great success of the Large language model (LLM), we use the Vicuna-7b model to combine the different captions on the same timestamp.

\begin{table}[]
    \centering
    \caption{Comparison with existing methods and different threshold of background information filtering.}
    \label{tab:Caption}
    \begin{tabular}{@{}c|c|c|c@{}}
    \toprule
    Method                    & Filter & CIDEr           & Meteor          \\ \midrule
    Flamingo                  & -      & 23.028          & 20.908          \\
    Blip-2                     & -      & 22.427          & 21.558          \\
    Blip                      & -      & 23.213          & 21.044          \\
    Blip + preceiver resampler & -      & \textbf{25.563} & \textbf{21.708} \\ \midrule
    Blip + preceiver resampler & 0.87   & 66.902          & 26.412          \\
    Blip + preceiver resampler & 0.875  & 70.116          & \textbf{26.832} \\
    Blip + preceiver resampler & 0.88   & \textbf{75.682} & 26.808          \\ \bottomrule
    \end{tabular}
    \end{table}

\section{Experiment}
\subsection{Experiment Settings}
Our model is trained on 8-GPU nodes with a batch size of 16. We use the AdamW~\cite{loshchilov2017decoupled} optimizer with a weight decay of 0.05. We use a cosine learning rate decay with a peak learning rate of 2e-4 and a linear warmup of 200 steps. We use the 16 seconds before and after the annotated timestamp and select the first frame per second. We randomly resized crop each frame to 256 * 256 and use horizontal flipping augmentation. 

\subsection{Ensemble Results}
In Table ~\ref{tab:Ensemble}, we compare the performance of different ensemble methods. We use grid search to select the weights of each model. The combination of Blip(ViT-Base) + Blip(ViT-Large)$\times$0.85 + Blip-2(ViT-Large)$\times$0.95 has the highest Meteor. However, due to the concern of over-fitting, we didn't select this method for the final submission.

\subsection{Efficacy of Perceiver Resampler and Filtering}
In our framework, we add perceiver resampler after the vision encoder. In Table\ref{tab:Caption}, all methods add a temporal embedding layer in the vision encoder to adapt for video input. Our framework outperforms Blip by +0.66$\%$ on the test set, which shows the effectiveness of perceiver resampler. Besides, compare with other methods our framework performs the best on both CIDEr and Meteor.

For localization, background information filtering is employed after the captions are generated. The results are shown in the bottom part of Table\ref{tab:Caption}. After adding filtering with threshold 0.875, the performance is improved by +6.124$\%$ in Meteor, which shows the effectiveness of filtering the background information. We also compare the performance of different filtering thresholds. The filtering threshold of 0.875 has the highest Meteor.

{\small
\bibliographystyle{unsrt}
\bibliography{egbib}
}

\end{document}